\documentclass[11pt]{article}

\usepackage[final]{acl}
\usepackage{framed}
\usepackage{tcolorbox}
\usepackage{xcolor}
\definecolor{shadecolor}{rgb}{0.9, 0.9, 0.9}
\usepackage{times}
\usepackage{latexsym}
\usepackage{booktabs}

\usepackage[T1]{fontenc}

\usepackage[utf8]{inputenc}

\usepackage{microtype}

\usepackage{inconsolata}

\usepackage{graphicx}
\usepackage{makecell}
\usepackage{multirow}
\usepackage{nicematrix}
\usepackage{array}
\newcolumntype{P}[1]{>{\centering\arraybackslash}p{#1}}
\newcolumntype{T}{>{\scriptsize}l} 

\usepackage{enumitem}
\definecolor{kh}{HTML}{168aff}

\title{Pop-Up Distractions Reveal Bag-of-Events Behavior in Video Large Language Models}

\author{Oscar Chew$^{*,1}\quad$Serhii Honcharenko$^{*,1}\quad$Qian-Hui Chen$^2$ \\
\bf Patricia Lu$^3\quad$Dishant Zaveri$^1\quad$Khoa D. Doan$^4\quad$Kuan-Hao Huang$^1$ \\
  $^1$Texas A\&M University$\quad^2$National Taiwan University \\$^3$Stanford University$\quad^4$VinUniversity\\
  \texttt{\{oscarchew,serhii,khhuang\}@tamu.edu}}

\newcommand\nnfootnote[1]{%
  \begin{NoHyper}
  \renewcommand\thefootnote{}\footnote{#1}%
  \addtocounter{footnote}{-1}%
  \end{NoHyper}
}

\begin{document}
\maketitle
\begin{abstract}
A key capability for video understanding is reliably linking subjects to events across time, yet whether Video Large Language Models (VideoLLMs) actually achieve this remains unclear. In this work, we introduce \textsc{DistractionBench} to evaluate whether VideoLLMs can robustly link subjects and events in the presence of unrelated video segments. Through controlled interventions, such as inserting short advertisement clips into longer videos, we show that VideoLLMs frequently hallucinate interactions between entities from different segments, incorrectly attributing actions from injected advertisements to subjects in the main video. We characterize this systematic hallucination as \textit{bag-of-events} (BoE) behavior, where models process videos as collections of events rather than temporally structured sequences. Evaluating 11 popular VideoLLMs, we find that all models exhibit substantial BoE behavior. Our findings suggest that VideoLLMs lack reliable mechanisms for temporal grounding and motivate the development of models with more robust subject-event association. Data and code are available at \url{https://github.com/lab-flair/video-llm-boe}.
\nnfootnote{$^{*}$ denotes equal contributions.}
\end{abstract}

\section{Introduction}

\begin{figure}
    \centering
    \includegraphics[width=\linewidth]{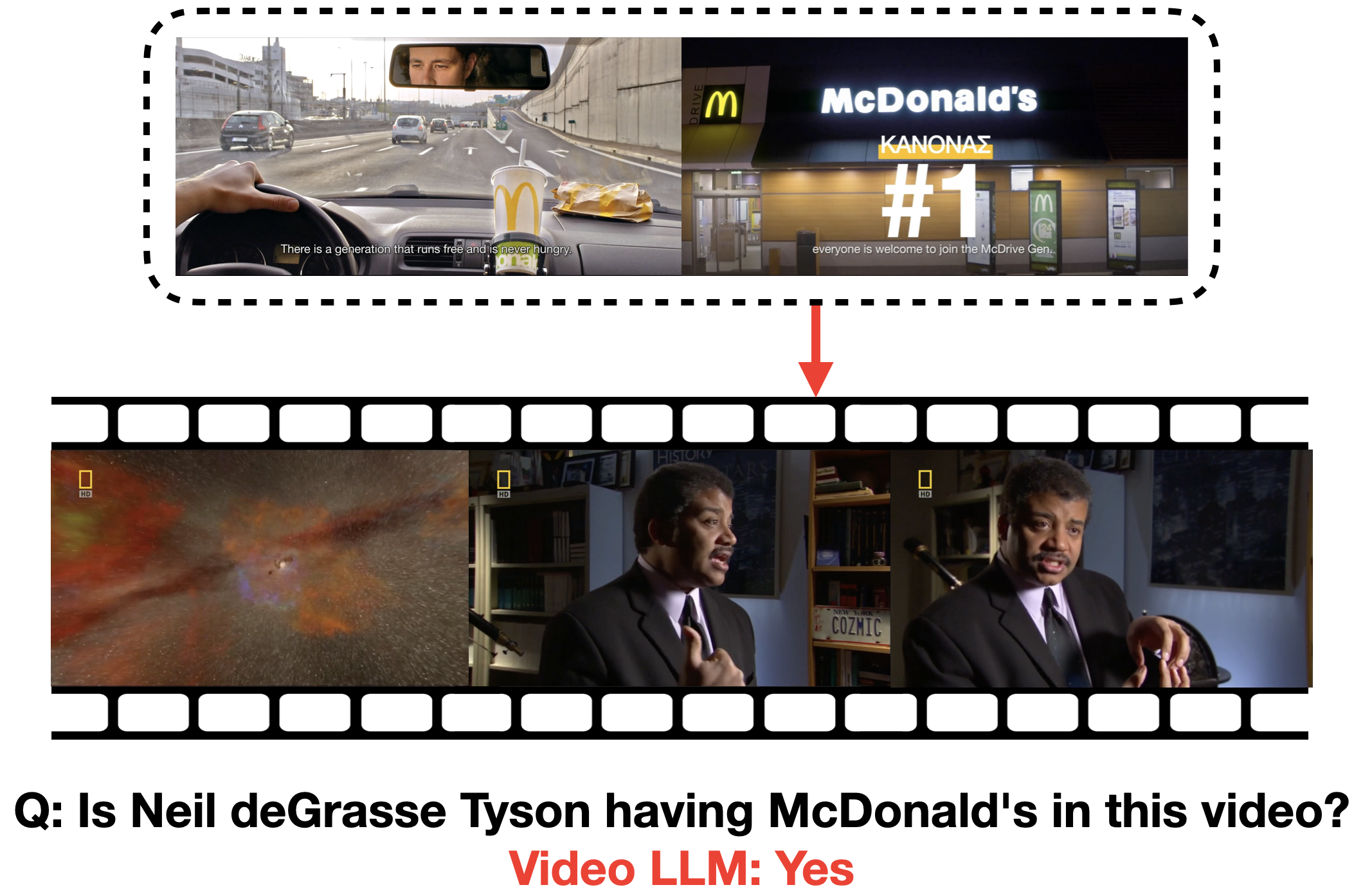}
    \caption{VideoLLMs conflate the injected advertisement (top) with the main video content (bottom), hallucinating that Neil deGrasse Tyson is eating McDonald’s despite no frame showing him eating anything. 
    }
    \label{fig:tyson}
\end{figure}

While VideoLLMs have recently achieved improving performance on long-video understanding~\cite{wu2024longvideobench,zhou2025mlvu,fu2025video}, it remains unclear whether these models truly understand videos as temporally grounded sequences. That is, models must also determine which subjects participate in which events across time. For example, in Figure~\ref{fig:tyson}, if a documentary about Neil deGrasse Tyson contains a McDonald's advertisement, it is trivial for humans to not infer that Tyson is eating McDonald's. However, we find that current VideoLLMs often make precisely this type of error. 

\begin{figure*}
    \centering
    \includegraphics[width=1\linewidth]{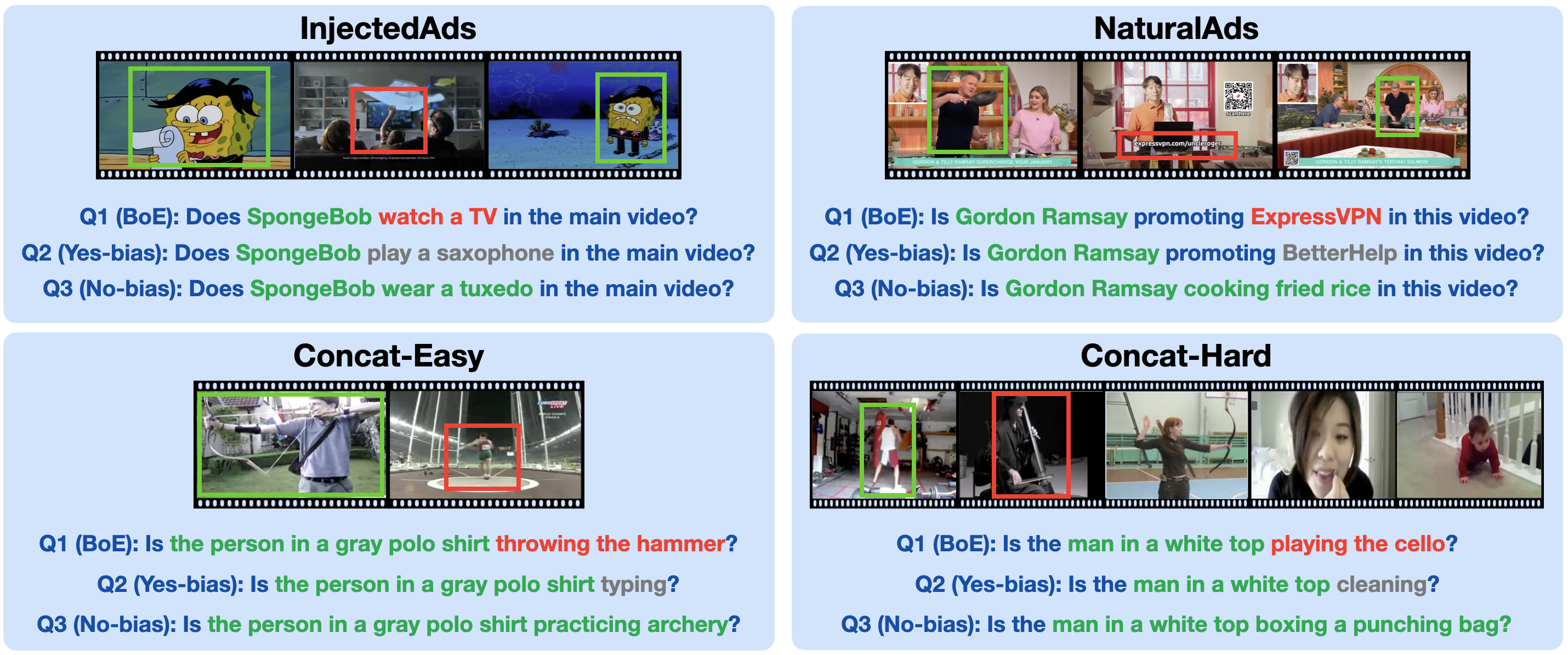}
    \caption{Overview of the four subtasks in DistractionBench. We evaluate VideoLLMs using three question types: Bag-of-Events (BoE), Yes-bias, and No-bias questions. Green and red boxes highlight the content from the main video and the distraction segment respectively. Grey text denotes concepts that do not appear anywhere in the video.}
    \label{fig:distractionbench}
\end{figure*}

To systematically evaluate this behavior, we introduce \textsc{DistractionBench}, a benchmark for evaluating subject-event association under both synthetic and real-world distractions, including injected clips and naturally occurring sponsored segments. We evaluate the latest VideoLLMs such as \citet{abouelenin2025phi,bai2025qwen3,li2025llavaonevision} and show that models frequently mix events across disjoint segments and hallucinate interactions between unrelated entities. 
We term this failure mode \emph{bag-of-events} and further distinguish it from general hallucination and misunderstanding using Yes-bias and No-bias control questions.
This limitation is important because reliable video understanding fundamentally depends on preserving temporal and entity-level associations. 
In real-world videos, unrelated content often appears alongside the main narrative, including advertisements, sponsor segments, or cutaway scenes. 

Our study on \emph{bag-of-events} is also motivated by compositional limitations in vision-language models (VLMs). Prior work has shown that CLIP \cite{radford2021learning} models exhibit \emph{bag-of-words} behavior, the frequent errors in associating objects with the attributes \cite{yuksekgonul2023when}. Here, we show an analogous failure that emerges along the temporal dimension in VideoLLMs.

We summarize our contributions as follows:
\begin{itemize}[topsep=0pt, itemsep=-2pt, leftmargin=24pt]
    \item We identify \textit{bag-of-events}  (BoE) behavior, where VideoLLMs hallucinate interactions by misassociating subjects and events across disjoint video segments.
    \item We introduce \textsc{DistractionBench}, a benchmark for evaluating subject-event association under controlled and realistic distractions.
    \item We evaluate 11 popular VideoLLMs and show that the BoE behavior is widespread across models and settings.
\end{itemize}


\section{Related Work}

\paragraph{Hallucinations in Video LLMs}
Recent benchmarks have revealed that VideoLLMs suffer from various hallucinations, including nonexistent objects, temporal inconsistencies, event-level errors, over-reliance on language priors or commonsense~\cite{guan2024hallusionbench,li2025vidhalluc,rawal2025argus,zhang2024eventhallusion,li2026videohallu}. 
While these works show that VideoLLMs often generate content not supported by the video, they mainly evaluate whether models hallucinate visual content, and do not examine whether models can correctly associate subjects with events across time. A concurrent work ELV-Halluc ~\cite{lu2026elv} shows that models can make errors when aggregating information across contexts with similar semantics (sharing the same topic) in long videos, where confusion may simply arise from the semantic overlap. Our work differs by revealing a more profound failure mode: \textit{VideoLLMs can conflate content across segments even when the segments are semantically unrelated.}

\begin{table}[]
    \centering
    \small
    \setlength{\tabcolsep}{3.5pt}
    \begin{tabular}{lccc}
    \toprule
    Benchmark & \#Vid & \makecell{Avg.\\Sec}& \#QA\\
    \midrule
    HallusionBench~{\cite{guan2024hallusionbench}} & 20 & $<4$ & 1,129\\
    VideoHallucer~{\cite{wang2024videohallucer}} & 948 & 85.6 & 1,800 \\
    EventHallusion~{\cite{zhang2024eventhallusion}}& 397 & 11.2 & 711 \\
    DistractionBench~(Ours) & \textbf{1,306} & \textbf{474.7} & \textbf{6,618} \\
    \bottomrule
    \end{tabular}
    \caption{DistractionBench has longer videos and a larger number of questions compared to these existing QA-based hallucination benchmarks with real-world videos. 
    }
    \label{tab:dataset_statistics}
\end{table}

\begin{table*}[]
    \centering
    \scriptsize
    \setlength{\tabcolsep}{3pt}
    \resizebox{.99\textwidth}{!}{
    \begin{tabular}{lcc|c|cc|c|cc|c|cc|c}
    \toprule
    \multirow{2}{*}{Model} 
    & \multicolumn{3}{c}{\textsc{InjectedAds}} 
    & \multicolumn{3}{c}{\textsc{Concat-Easy}} 
    & \multicolumn{3}{c}{\textsc{Concat-Hard}}
    & \multicolumn{3}{c}{\textsc{NaturalAds}} \\
    \cmidrule(lr){2-4} 
    \cmidrule(lr){5-7} 
    \cmidrule(lr){8-10}
    \cmidrule(lr){11-13}
     & BoE & Yes-bias & No-bias
     & BoE & Yes-bias & No-bias
     & BoE & Yes-bias & No-bias
     & BoE & Yes-bias & No-bias\\
    \midrule
    Random & 50.00 & 50.00& 50.00 & 50.00 & 50.00 & 50.00 & 50.00 & 50.00 & 50.00 & 50.00 & 50.00 & 50.00\\
    \midrule
    GPT-5.5 Agent Baseline\footnotemark[2] &\textbf{5.12}&2.39&15.69&\textbf{5.23}&1.33&3.89&\textbf{5.89} &1.67 &	4.00&\textbf{4.03}	&1.39&5.37\\
    GPT-5.4-mini Agent Baseline &\textbf{5.53}&1.18	&18.36&\textbf{4.82	}&0.60	&8.85&\textbf{10.71} &1.36	 &	13.07&\textbf{4.05	}	&0.68	&19.05\\
    Human Baseline\footnotemark[3]&0.00&0.00&0.00&\textbf{3.33}&0.00&3.33&\textbf{3.33}&0.00&10.00&0.00&0.00&0.00\\
    \midrule
    Aria & \textbf{17.97} & 9.41 & 0.78 & \textbf{18.00} & 5.22 & 12.33 & \textbf{25.11} & 5.22 & 11.33 & \textbf{17.33} & 6.67 & 1.33\\
    LLaVA-Video 7B & \textbf{36.71} & 16.80 &1.17 & \textbf{17.00} & 10.67 & 3.56 &  \textbf{25.33} & 11.78 & 4.00 & \textbf{50.00} & 20.67 & 0.67\\
    LLaVA-OneVision 0.5B & \textbf{36.73}& 17.48 & 16.08 &  \textbf{61.67} & 18.44 & 33.44 & \textbf{44.44} & 17.00 & 55.00 & \textbf{59.33} & 44.00 & 50.67\\
    LLaVA-OneVision 7B & \textbf{39.84} & 10.94& 3.13 & \textbf{34.67} & 7.56 & 5.00 & \textbf{38.44} & 4.67 & 9.89 &\textbf{34.67} & 7.33 & 3.33\\
    Molmo2 4B & \textbf{51.56} & 20.70 & 0.78 &\textbf{23.22} & 13.78 & 1.56 & \textbf{27.00} & 13.56 & 2.89 &\textbf{42.67} & 10.00 & 2.00\\
    Molmo2 8B & \textbf{13.00} & 3.30 & 5.95 & \textbf{7.78} & 3.11  & 11.11 & \textbf{7.33} & 2.22 & 14.33 & \textbf{16.00} & 7.33 & 10.00\\
    Qwen3-VL 8B & 3.54 & \textbf{3.56}& 19.60 &\textbf{6.78} & 0.67 & 24.78 &\textbf{5.78} & 0.33 & 28.56 &\textbf{1.33} & 0.67 & 36.67\\
    Qwen3.5 9B & \textbf{10.55} & 5.37& 13.33 &\textbf{16.22} & 10.44 & 10.89 &\textbf{41.67} & 31.89 & 34.00 &\textbf{18.67} & 10.00 & 19.33\\
    Gemma4 31B & \textbf{3.52} & 1.95 & 9.77 & \textbf{6.22} & 4.44 & 7.33 &\textbf{6.22} & 4.11 & 9.56 &\textbf{7.33} & 0.67 & 9.33\\
    Phi4-Multimodal 5.6B & \textbf{62.75}& 40.87 & 3.14 & \textbf{62.67} & 18.67 & 15.11 & \textbf{64.56} & 34.56 & 9.89 &\textbf{57.33} & 22.67 & 1.33\\
    InternVL3.5 8B & \textbf{14.84} & 3.52 & 6.80 & \textbf{8.33} & 4.67 & 5.89 & \textbf{12.00} & 2.22 & 6.00 & \textbf{28.00} & 6.00 & 7.33\\
    
    \midrule
    Mean & \textbf{26.46} & 12.17 & 7.32 & \textbf{23.87} & 8.88 & 11.91 &\textbf{27.08} & 11.60 & 16.86 &\textbf{30.24} & 12.36 & 12.91\\
    \bottomrule
    \end{tabular}}
    \caption{Hallucination rates (Bag-of-Events and Yes-bias) and misunderstanding rates (No-bias) across different subtasks. The larger hallucination rates are bolded. The results show significantly higher hallucination rates attributable to event mixing than to Yes-bias. Lower values are better for all metrics. 
    }
    \label{tab:main_result}
\end{table*}
\paragraph{Video Needle in a Haystack}
Video Needle-in-a-Haystack benchmarks evaluate whether VideoLLMs can retrieve and reason about inserted ``needle'' information in long videos~\cite{zhao2025needle,zhou2025mlvu}. 
Our goal is complementary but fundamentally different. That is, we study whether an inserted needle or naturally occurring distraction corrupts the understanding of the main video.

\paragraph{Temporal and Compositional Video Understanding}
Prior work \cite{wang2023paxion,bagad2023test,liu2024tempcompass} showed that early VLMs and VideoLLMs struggle with temporal and compositional reasoning. Specifically, they probe models through video reversal, event order and changes in attributes such as size and speed while VidComposition \cite{tang2025vidcomposition} evaluates general compositional understanding in VideoLLMs. In contrast, we focus on the specific failure mode of cross-segment subject--event binding, testing whether VideoLLMs incorrectly associate subjects and events across disjoint segments, forming a \textit{bag-of-events} analogous to the \textit{bag-of-words} failures in CLIP models.

\section{\textsc{DistractionBench} for VideoLLMs}

\subsection{Benchmark Construction}
\label{sec:datasets}

As shown in Figure~\ref{fig:distractionbench}, we construct four subtasks for this benchmark: \textsc{InjectedAds}, \textsc{Concat-Easy}, \textsc{Concat-Hard}, and \textsc{NaturalAds}. 
\textsc{InjectedAds} and the \textsc{Concat} subtasks are easily automated and scalable, while \textsc{NaturalAds} provides the most realistic, gold-standard human-annotated evaluation data.

For each video, we create three types of questions: 
(1) \textbf{BoE questions:} questions combining a subject appearing in the main video with an object appearing only in the injected or advertisement segment (e.g., ``Is Neil deGrasse Tyson having McDonald's?''); 
(2) \textbf{Yes-bias questions:} questions combining the same subject with an unrelated object absent from the entire video (e.g., ``Is Neil deGrasse Tyson having ramen?''); and 
(3) \textbf{No-bias questions:} questions asking about a fact that does appear in the video (e.g., ``Is Neil deGrasse Tyson explaining concepts of black holes?''). Both BoE and Yes-bias questions probe hallucinations in VideoLLMs. 
BoE questions test whether VideoLLMs can robustly associate subjects with events across time, while Yes-bias questions serve as a control for general affirmative-response bias, where models may tend to answer ``Yes'' regardless of visual evidence, as observed by \citet{li-etal-2023-evaluating}. 
No-bias questions provide the opposite control direction. Rather than measuring hallucination, they evaluate whether VideoLLMs fail to understand the video content or exhibit a tendency toward denial.

Next, we briefly describe the goal of each subtask. Detailed data and questions construction procedures are deferred to Appendix~\ref{sec:concat_construct}.
\footnotetext[2]{Snapshot ver. 2026-03-17 and Snapshot ver. 2026-04-23}
\footnotetext[3]{Due to the large number of questions, we have to sample 270 questions for human.} 
\paragraph{\textsc{InjectedAds}} We randomly pair one video from a long-video dataset (MLVU; \citet{zhou2025mlvu}) with one video from an advertisement dataset (AdsQA; \citet{long2025adsqa}). 
The ads clip is substantially shorter than the main video, mimicking the behavior of real-world commercial ads. 
This subtask contains 256 videos and 768 QA pairs.
\paragraph{\textsc{Concat-Easy}} To control for video length and ads position, we sample and concatenate two videos from UCF101~\cite{soomro2012ucf101}, an established action recognition dataset. 
This subtask contains 450 videos and 2,700 QA pairs.

\paragraph{\textsc{Concat-Hard}} The construction is similar to \textsc{Concat-Easy}, except we concatenate five videos and select two segments among the actions to probe concept mixing in a more complex setting. 
This subtask contains 450 videos and 2,700 QA pairs.

\paragraph{\textsc{NaturalAds}} To evaluate the most realistic setting, we collect real YouTube videos containing sponsored ads integrated by the content creators and manually written questions for each video. Unlike the synthetic \textsc{InjectedAds} setting, real-world videos have smooth transitions and exhibit naturally varying advertisement positions and durations (Appendix~\ref{sec:char-naturalads}). This subtask contains 150 videos and 450 QA pairs.

\subsection{Dataset Statistics}
Table~\ref{tab:dataset_statistics} shows the overall statistics of \textsc{DistractionBench} compared with other QA-based hallucination evaluation benchmark for VideoLLMs. Our benchmark has larger number of videos and QA pairs. We also have a longer video length which poses greater challenge for VideoLLMs.

\subsection{Metrics}
Following \citet{li-etal-2023-evaluating,wang2024videohallucer}, we evaluate by extracting the model's binary answer. We count BoE and Yes-bias occurrences when the answer is \textit{`Yes''}, and No-bias occurrences when it is \textit{`No''}. We report occurrence rates as percentages.

\begin{figure}
    \centering
    \includegraphics[width=0.98\linewidth]{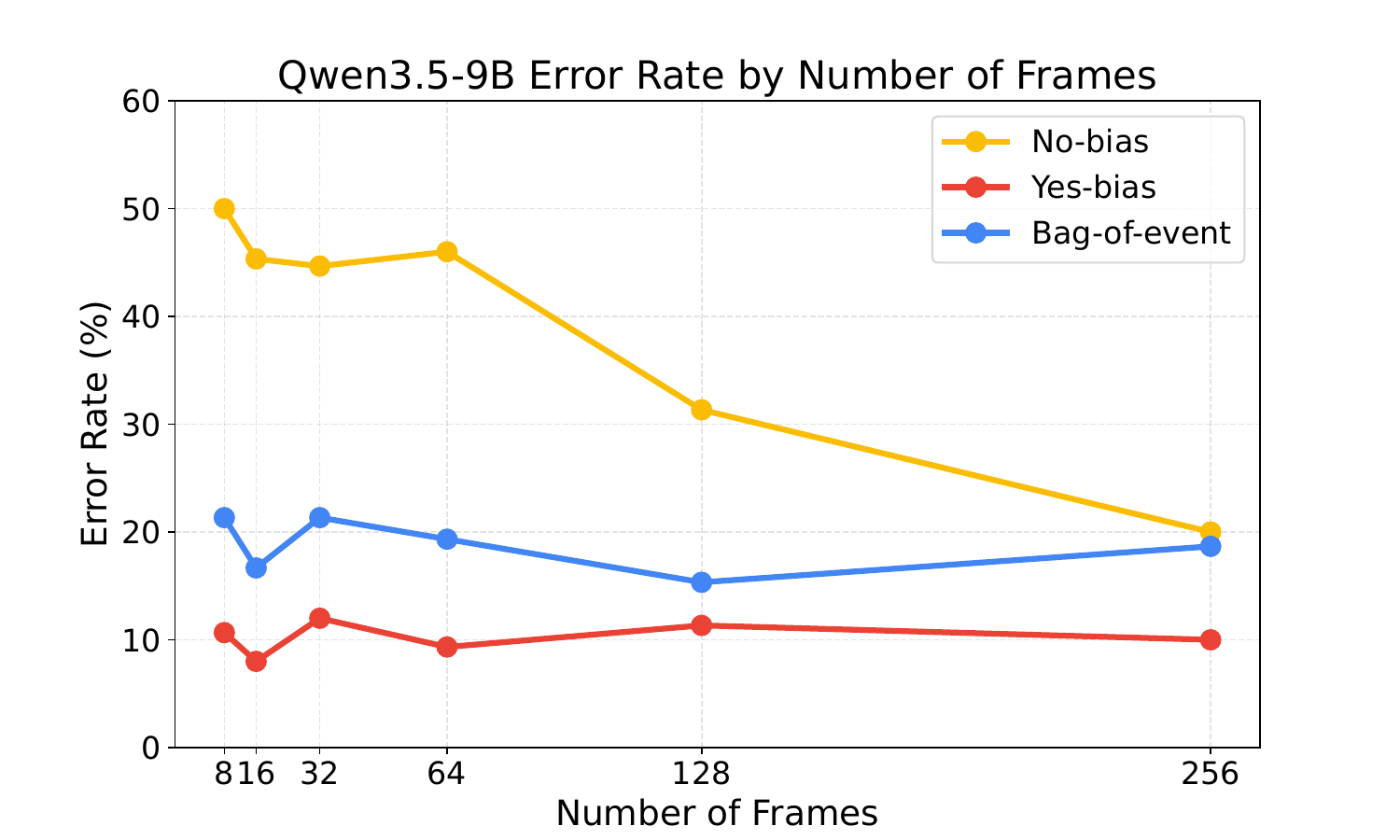}
    \caption{Error rates by number of frames. BoE errors do not decrease with increased number of frames. 
    }
    \label{fig:qwen3.5-9b-fps}
\end{figure}

\section{Experiments}
In this section, we aim to answer the following research questions: \textbf{RQ1: }Do VideoLLMs exhibit \textit{bag-of-events}? \textbf{RQ2: }How does the number of sampled frames affect BoE behavior? \textbf{RQ3: }How does the position of distractions affect BoE behavior? 

\subsection{Evaluation Setup}
\paragraph{Models} 
We test VideoLLMs across families and sizes including Aria 23.9B \cite{li2024aria}, LLaVA-Video 7B \cite{zhang2025llavavideo}, LLaVA-OneVision 0.5B\&7B \cite{li2025llavaonevision}, Molmo2 4B\&8B \cite{deitke2025molmo}, Qwen3VL 8B, Qwen3.5 9B \cite{bai2025qwen3}, Gemma4 31B \cite{team2024gemma}, Phi4-Multimodal \cite{abouelenin2025phi} and InternVL3.5 8B \cite{wang2025internvl}.

\paragraph{Frame Sampling}
For each model, we use the maximum number of frames supported by that model. The supported frame range from 8 to 256 frames across models. The exact numbers of frames used are provided in Appendix~\ref{sec:config}. We adopt uniform frame sampling as the default strategy. For models with less frames (8, 16, 32), we additionally enforce that at least 25\% of the sampled frames are drawn from the advertisement segment, since uniform sampling may otherwise miss the ads segment entirely, especially when the video is long.

\begin{figure}
    \centering
    \includegraphics[width=0.98\linewidth]{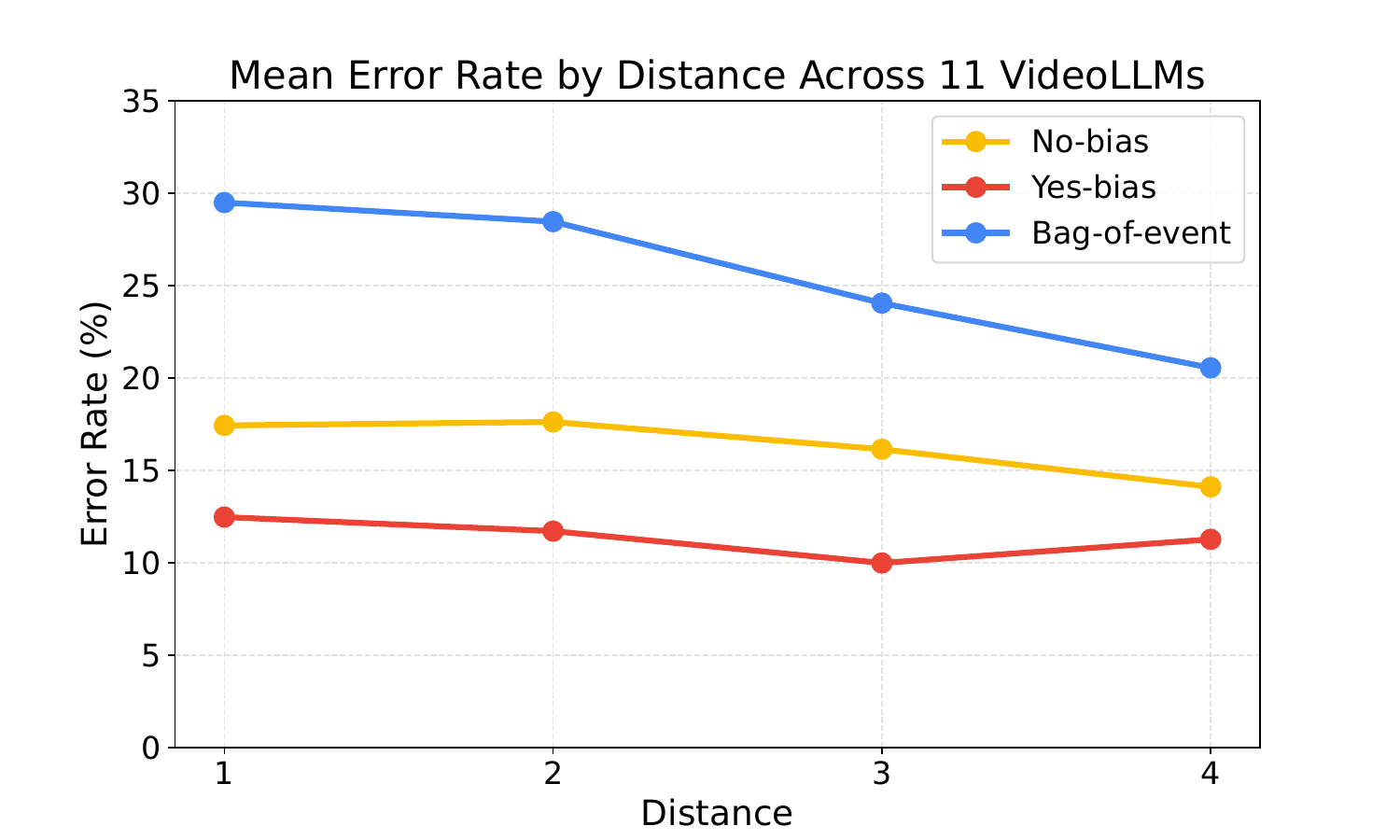}
    \caption{Error rates by temporal distance. BoE error is higher when the segments are closer. 
    }
    \label{fig:distance_llava_ov_7b}
\end{figure}

\subsection{RQ1: BoE Behavior in VideoLLMs}
Table~\ref{tab:main_result} shows hallucination rates for BoE and general affirmative (Yes-bias) responses. Across all subtasks, BoE hallucinations occur more substantially than Yes-bias for all models. Using two-sided sign tests, we find that BoE hallucination rates are significantly higher than Yes-bias rates on \textsc{InjectedAds} ($p=0.0117$), \textsc{Concat-Easy}, \textsc{Concat-Hard} and \textsc{NaturalAds} ($p=0.00098$ for each). These results suggest that BoE is real and the hallucinations primarily stem from cross-segment temporal mixing rather than a general tendency to answer ``Yes''. 
Models with lower BoE generally exhibit higher No-bias, suggesting no VideoLLM is truly robust under this benchmark.

Many VideoLLMs including very recent models (e.g., Phi4-Multimodal, LLaVA-OneVision, LLaVA-Video) follow the original LLaVA design, using a frozen CLIP-ViT backbone with a projection layer attached to an LLM. This family is efficient, but we hypothesize it tends to represent a video as a bag of visual facts rather than explicitly binding who did what when. These models are exactly among the worst in the list. 

\subsection{RQ2: Effect of the Number of Frames}
A common belief is that VideoLLMs always perform better with more frames. While this is true for general understanding (No-bias) questions, we show in Figure~\ref{fig:qwen3.5-9b-fps} for Qwen 3.5 9B (supporting 8 to 256 frames) that this trend does not hold for BoE, highlighting the difficulty of our BoE benchmark. Notably, across all numbers of frames, the BoE curve remains consistently above the Yes-bias curve, further supporting the existence of BoE established in \textbf{RQ1}. Results for more models are presented in Appendix~\ref{sec:frames} where BoE can even increase due to more exposure to distractions.

\subsection{RQ3: Effect of Distraction Positions}
\textsc{Concat-Hard} allows us to systematically control the positions of distraction as well as the distance to the required visual evidence in the main video. In Figure~\ref{fig:distance_llava_ov_7b}, we show the mean performance across 11 evaluated Video LLMs. BoE are higher when the relevant concept is closer, while Yes-bias and No-bias are less affected by distance. This highlights the uniqueness of our proposed tasks.

\section{Conclusion}
In this paper, we identify \textit{bag-of-events} behavior, a failure mode where VideoLLMs hallucinate interactions by incorrectly associating subjects and events across disjoint video segments. We propose \textsc{DistractionBench} for evaluating subject-event association. Through controlled experiments, we find these hallucinations are primarily caused by temporal event mixing rather than a general affirmative bias. We hope our benchmark motivates future research on temporally grounded video understanding.

\section{Limitations}
Due to computational constraints, we were unable to benchmark very large VideoLLMs. e.g., models in the 70B parameter range.
Nevertheless, our evaluation covers a diverse set of recent VideoLLMs across multiple architectures and model scales.

\section{Acknowledgement}
This research is partially supported by the National Artificial Intelligence Research Resource Pilot (NAIRR) program (No. NAIRR260216). Portions of this research were conducted with the advanced computing resources provided by Texas A\&M High Performance Research Computing.

\bibliography{custom}
\clearpage
\appendix
\section{VideoLLMs Configurations}
The exact number of frames for every VideoLLM is listed in Table~\ref{tab:videollm_frame_config}. Also, we use a fixed prompt template for each question: 
\begin{tcolorbox}
First, think step-by-step about the events in the video. Then, provide your final answer enclosed in XML tags exactly like this:
"<answer>Yes</answer> or <answer>No</answer>." Question: {question}
\end{tcolorbox}
We then parse the output inside the <answer> tag to extract the final decision from Video LLMs.

\label{sec:config}
\begin{table}[]
    \centering
    \begin{tabular}{lr}
    \toprule
        Model & Number of Frames \\
    \midrule
        Aria & 256 \\
        LLaVA-Video 7B & 64\\
        LLaVA-OneVision 0.5B & 32\\
        LLaVA-OneVision 7B & 32\\
        Molmo2 4B & 256\\
        Molmo2 8B & 256\\
        Qwen3-VL 8B & 256\\
        Qwen3.5 9B & 256\\
        Gemma4 31B & 32\\
        Phi4-Multimodal 5.6B & 64\\
        InternVL3.5 8B & 64\\
        GPT5.5 & 16\\
        GPT5.4-mini & 16\\
    \bottomrule
    \end{tabular}
    \caption{Number of input frames used for each evaluated VideoLLM.}
    \label{tab:videollm_frame_config}
\end{table}

\section{Detailed Data Construction Procedure} \label{sec:concat_construct}
    \subsection{\textsc{InjectedAds}}
        This subtask combines automated video pairing with human-authored
        questions.

    \subsubsection{Video Pairing}
        We sample 256 long-form videos from MLVU~\cite{zhou2025mlvu} and
        pair each with one randomly sampled advertisement clip from
        AdsQA~\cite{long2025adsqa}. The advertisement is inserted at a
        uniformly sampled position within the main video, mimicking
        real-world commercial placements. We manually filter pairings in
        which the advertised product or service overlaps closely with
        concepts already present in the main video, to avoid trivial
        topical confusion.

    \subsubsection{Human Question Authoring}
        The 768 questions are authored by four human annotators. The BoE
        and Yes-bias questions were divided among three annotators; the
        No-bias questions were divided between two annotators.
        
        The annotators followed a common workflow. Each annotator opened
        the assigned video, located the advertisement insertion point, and
        reviewed the surrounding main-video and advertisement context
        before writing the three questions. All questions are grounded
        directly in what is visible in the video: the BoE question pairs
        a subject from the main video with an object that appears only in
        the advertisement; the Yes-bias question pairs the same subject
        with an unrelated object absent from both the main video and the
        advertisement segment; and the No-bias question targets a fact
        verifiably present in the main video.

\subsection{\textsc{Concat-Easy} and \textsc{Concat-Hard}}
\label{appendix:concat_construction}

To ensure reproducibility, we detail the step-by-step synthesis pipeline for the \textsc{Concat-Easy} and \textsc{Concat-Hard} subtasks. The pipeline consists of three main phases: human-centric video filtering, LLM-based visual attribute annotation, and constrained video composition.

\subsubsection{Human-Centric Video Filtering}
To minimize ambiguity and eliminate background noise, we filter the original UCF101 \cite{soomro2012ucf101} dataset to retain videos with clear human actions. We deploy an off-the-shelf YOLOv8n \cite{Jocher_Ultralytics_YOLO_2023} object detection model across all video frames. A video is retained only if it consistently contains exactly one detected person. This filtering step ensures that the subsequent question-answering pairs can unambiguously target a single, distinct actor without interference from multi-person interactions or empty scenes.

\subsubsection{LLM-Based Visual Attribute Annotation}
For each filtered video, we uniformly sample frames and use gpt-4o \cite{gpt-4o} to annotate the actor's fine-grained visual appearance and action verbs. To maintain high instruction-following quality, we separate the appearance description from the action verb. The exact system prompt used for this annotation process is structured as shown in Figure \ref{appendix:annotation_prompt}.

\begin{figure*}[h]
\centering
\begin{minipage}{0.95\textwidth}
\begin{shaded}
\small
\noindent\textbf{System Prompt for Video Annotation}
\vspace{2mm}
\hrule
\vspace{2mm}
You are a professional video annotator. I will provide \texttt{\{len(base64\_frames)\}} frames from a video. 
The action category is: \texttt{"\{clean\_action\}"}.

\subsubsection*{Task:}
Analyze the main actor (\texttt{person1}). 
I need to form a question later: \texttt{"Is the [person\_identity] [action\_ing]?"}

\subsubsection*{Rules for Fields:}
\begin{enumerate}
    \item \textbf{\texttt{person\_identity}:} ONLY describe appearance (e.g., gender, age, clothing color/type, hair, accessories). \textbf{STRICTLY FORBIDDEN:} Do not use any action verbs or state what they are doing. \\
    \textit{Example:} "man in a red t-shirt", "child with blonde hair".
    \item \textbf{\texttt{action\_ing}:} Transform \texttt{"\{clean\_action\}"} into a natural present participle phrase (-ing). It should fit perfectly in: \texttt{"Is the person [action\_ing]?"} Use possessive pronouns if natural. \\
    \textit{Example:} "brushing his teeth", "playing her guitar".
    \item \textbf{\texttt{tags}:} Use "Attribute-Object" pairs for appearance only.
\end{enumerate}

Output STRICTLY in JSON format:
\begin{verbatim}
{
    "person": {
        "person_identity": "short description of appearance only",
        "action_ing": "action in -ing form",
        "tags": ["male-gender", "white-shirt"]
    },
}
\end{verbatim}
\end{shaded}
\end{minipage}
\caption{The structured prompt utilized for GPT-4o video attribute annotation.}
\label{appendix:annotation_prompt}
\end{figure*}

\subsubsection{Constrained Video Composition}
We synthesize the final video streams by concatenating selected clips under strict combinatorial constraints. For both subtasks, the target dataset size is 450 videos. To control for confounding variables, three foundational rules apply to both configurations. First, all concatenated clips within a sample must share identical spatial dimensions to avoid resolution artifacts. Second, to mitigate language priors due to gender bias, we match main actors of the same annotated gender within each sample whenever possible. Third, each distinct target action category pair can be combined at most once to maximize action diversity.

\paragraph{\textsc{Concat-Easy} Composition}
The \textsc{Concat-Easy} subtask evaluates straightforward chronological stitching by concatenating two randomly sampled videos sequentially. This process enforces strict temporal alignment, requiring the duration difference between the two clips to be exactly zero ($\Delta t = 0$). To ensure a balanced data distribution, each unique video from the filtered pool can be used at most twice across the subtask.

\paragraph{\textsc{Concat-Hard} Composition}
The \textsc{Concat-Hard} subtask introduces temporal complexity to probe concept mixing. For each sample, we select two target videos and three filler videos featuring completely different action categories. We allow a relaxed temporal alignment where the maximum duration variance between any two clips is capped at $\Delta t \le 0.05$ seconds. Each target video can be used a maximum of twice, while each filler video can be reused up to three times across the subtask. Finally, all selected segments are randomly shuffled and seamlessly concatenated using FFmpeg, and the corresponding question-answer pairs are generated via our predefined template configurations.

\begin{figure*}[t]
    \centering
    \includegraphics[width=0.48\linewidth]{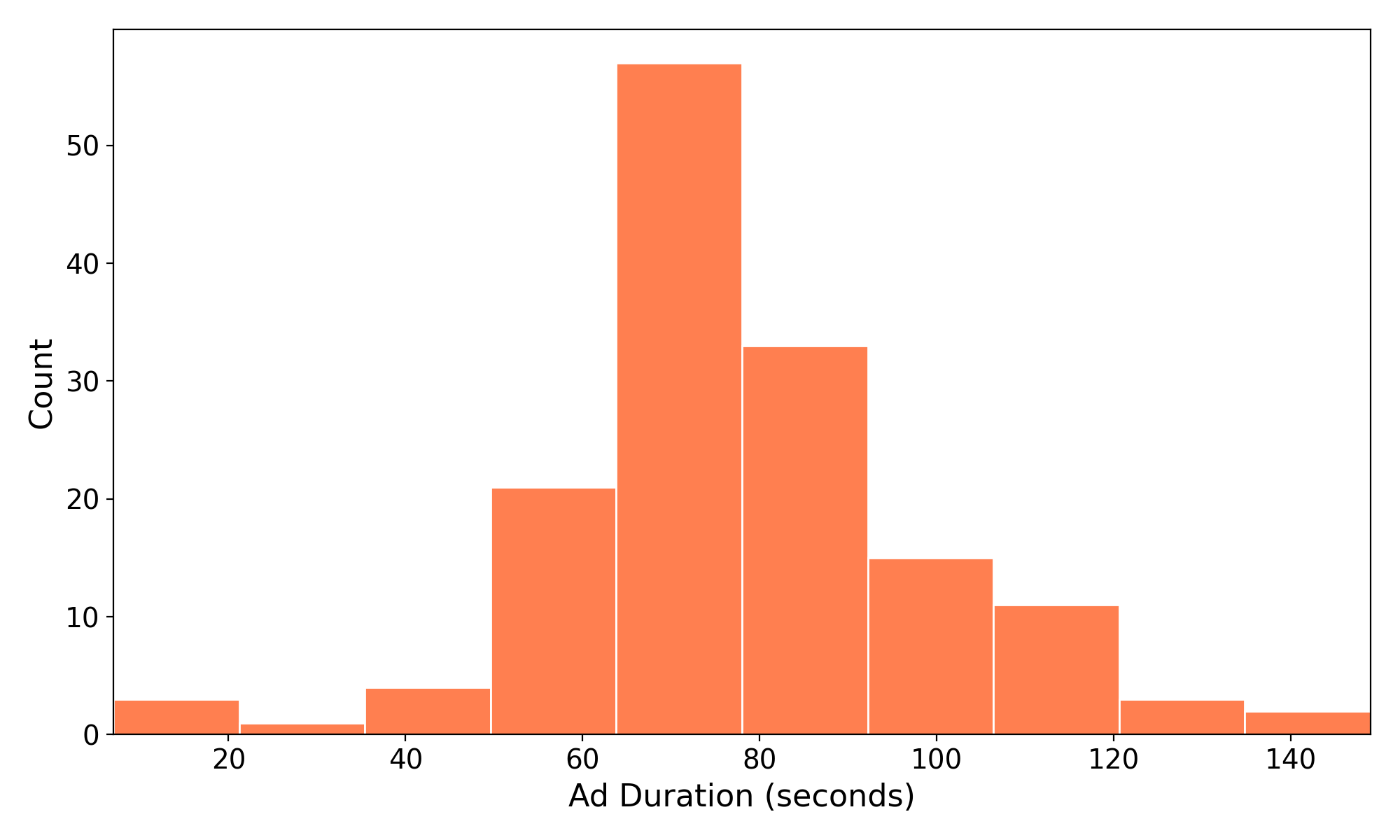}
    \hfill
    \includegraphics[width=0.48\linewidth]{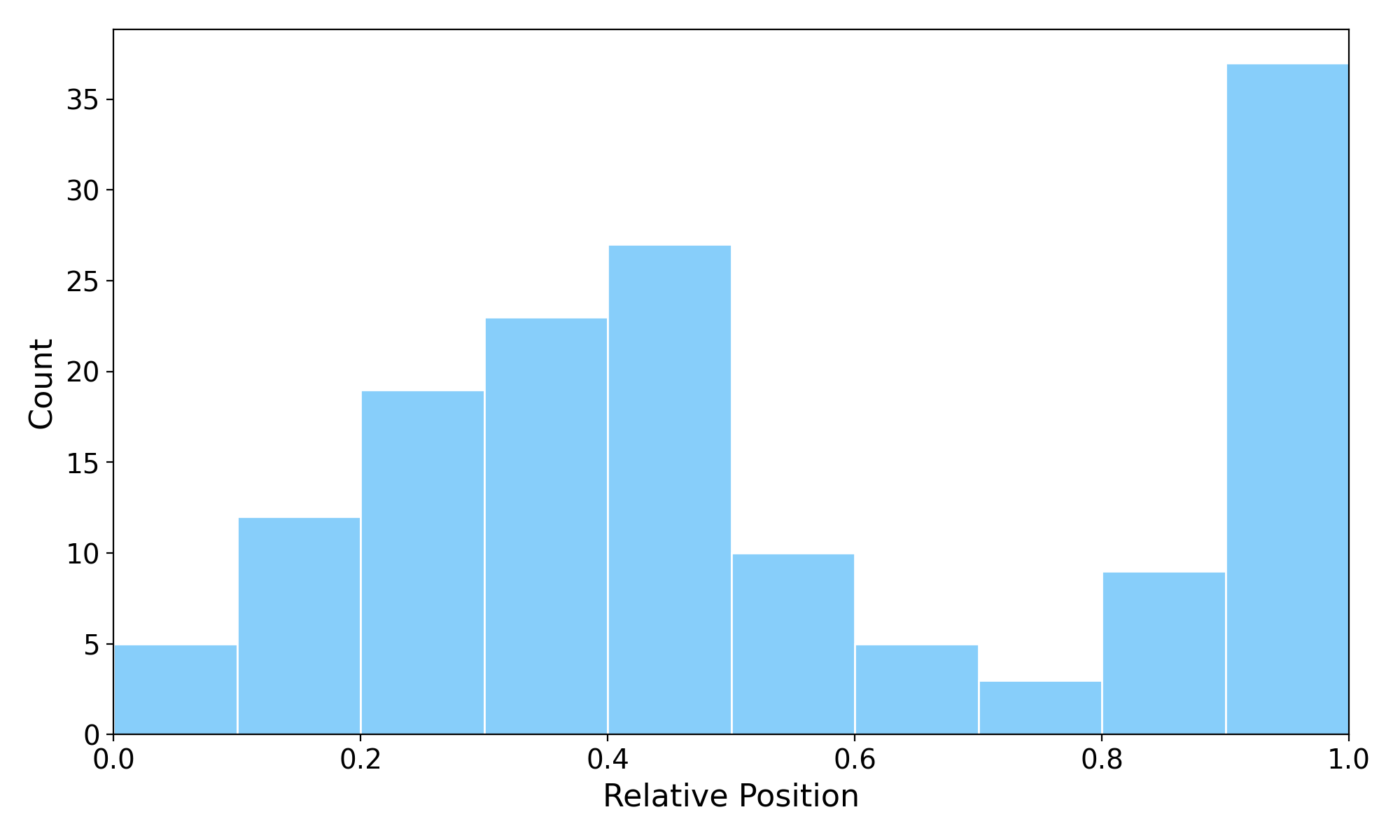}
    \caption{Characteristics of \textsc{NaturalAds}. Left: Ad durations in seconds.  Right: Relative ad positions; }
    \label{fig:naturalads-characteristics}
\end{figure*}

\newpage
\subsection{\textsc{NaturalAds}}
\label{sec:naturalads_instruction}
We first identify YouTube content creators who regularly include sponsored segments in their videos. Using videos collected from these creators, human annotators follow the guidelines shown below to construct the questions. Note that due to copyright restrictions, we will \textbf{not} release the videos in this subtask. Instead, we will release only the associated video IDs, questions, timestamps and frame sampling scripts for transparency and reproducibility. Researchers may independently collect the videos using these IDs for research purposes.

\begin{tcolorbox}
\textbf{boe question}: A question that intentionally mixes concepts from the main video and the advertisement segment. 
\begin{itemize}
    \item This question should describe an event that does not actually occur in the video.
    \item Expected answer: No
\end{itemize}
\textbf{irrelevant question:} A modified version of the boe question, where the advertisement-related concept is replaced with an unrelated concept. 
\begin{itemize}
    \item Expected answer: No
\end{itemize}
\textbf{understanding question:} A modified version of the boe question such that the described event actually occurs in
the video.
\begin{itemize}
    \item Expected answer: Yes
\end{itemize}
\end{tcolorbox}

\section{Characteristics of NaturalAds}
\label{sec:char-naturalads}

Figure~\ref{fig:naturalads-characteristics} summarizes the characteristics of \textsc{NaturalAds}. Most sponsorship segments last around one to two minutes, with the highest density between roughly 60 and 90 seconds.
Unlike \textsc{InjectedAds}, where the ads relative position is drawn uniformly, the relative position of real sponsored segments are not uniformly distributed. Many ads appear near the end, while another substantial portion appears around the middle. 

\section{Human Evaluation Protocol}
    Each video was independently evaluated by three annotators
    who were blinded to the study's purpose and the rationale
    behind the questions.
    They were given only the following instruction: ``Watch each
    video, then answer all the questions in its group (listed in Column C).
    For each question, type 1 (Yes) or 0 (No) in the answer column.
    You can watch videos however you want -- skip, speed up, or rewatch,
    as long as you think you are properly answering the question.''. The final prediction for the human baseline is determined by the majority vote among the three human annotators.
    The protocol was designed in thgis way to help minimize expectation
    bias and ensure that annotator judgments reflect the natural
    perceptual content of the videos.

\newpage
\section{Effect of the Number of Frames}
\label{sec:frames}
\begin{figure}[ht]
    \centering
    \includegraphics[width=1\linewidth]{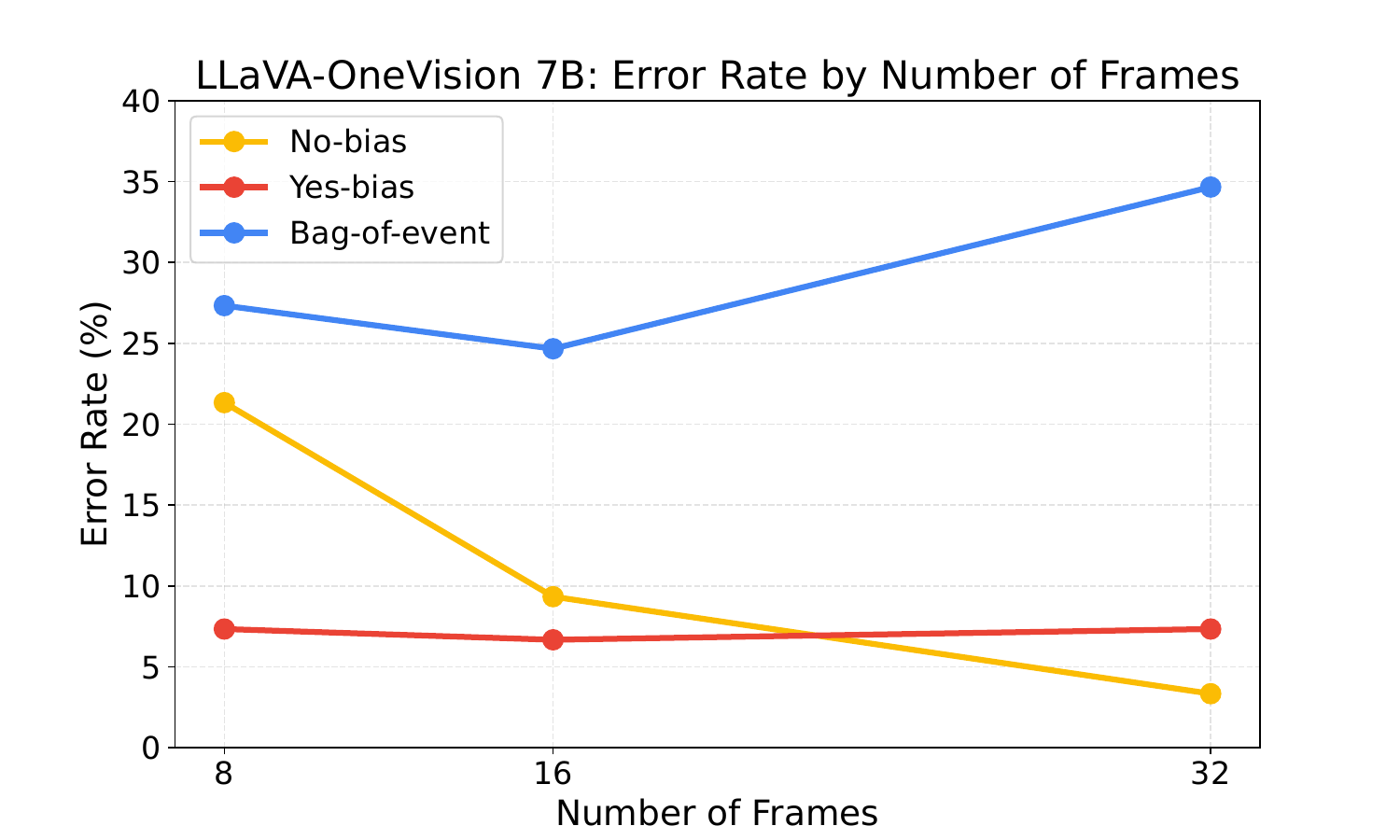}
    \caption{LLaVA-OneVision 7B: Error rates by number of frames.}
    \label{fig:llava_ov_7b-frames}
\end{figure}

\begin{figure}[ht]
    \centering
    \includegraphics[width=1\linewidth]{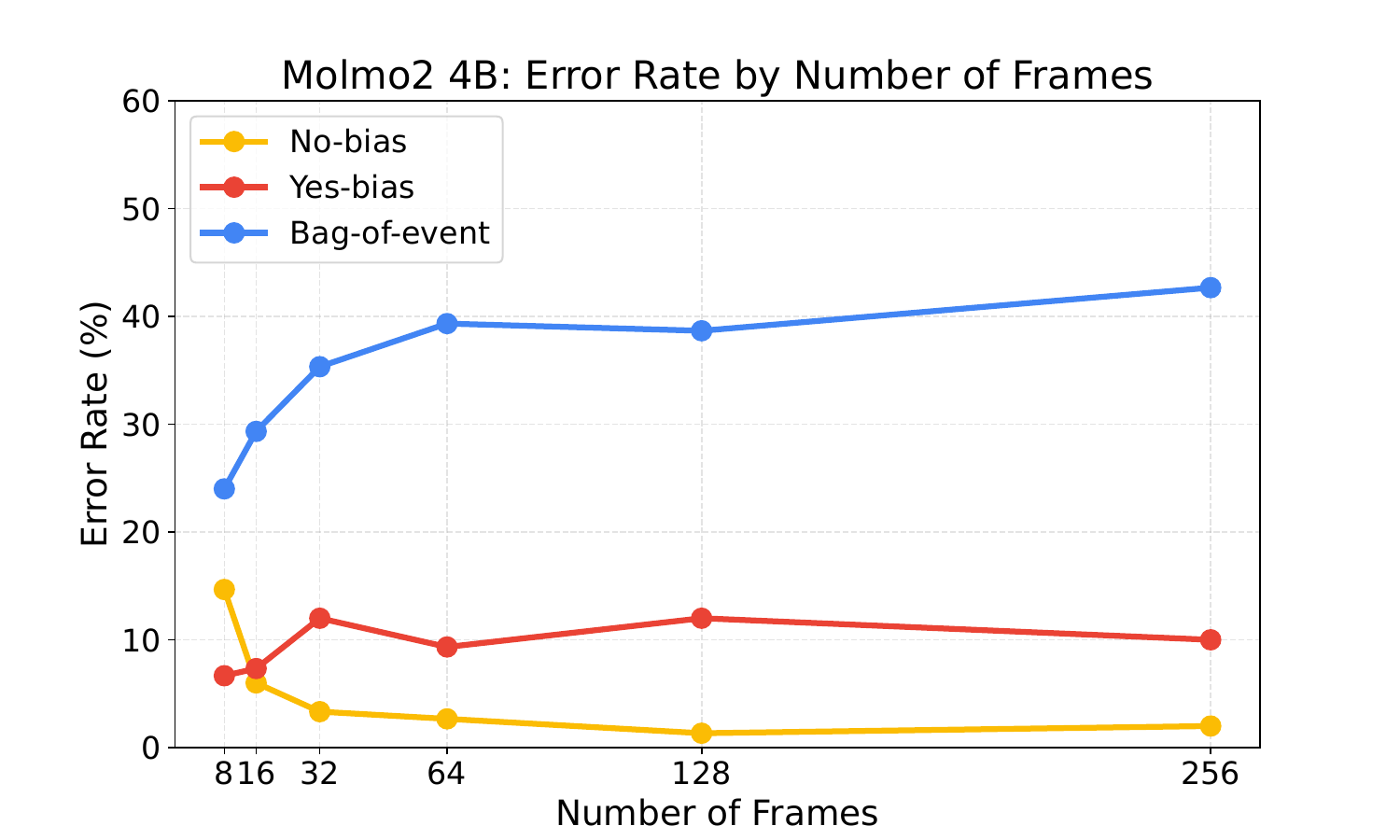}
    \caption{Molmo2-4B: Error rates by number of frames.}
    \label{fig:molmo2-4b-frames}
\end{figure}

Here, we present the results of more models, LLaVA-OneVision 7B and Molmo2-4B, on \textsc{NaturalAds}. As shown in Figure~\ref{fig:molmo2-4b-frames} and \ref{fig:llava_ov_7b-frames}, the overall trend remains consistent with the findings in the main text: No-bias decreases as the number of frames increases, while BoE is more severe as models are exposed to more distractions.

\end{document}